\documentclass[a4paper,conference]{IEEEtran}
\ifCLASSINFOpdf
\else
\fi
\usepackage{multirow}
\usepackage{algorithm}
\usepackage{algorithmic}
\usepackage{subfigure}
\usepackage{mathtools} 
\usepackage{threeparttable}

\usepackage{amsmath}
\begin{document}
%
\title{Is the Meta-Learning Idea Able to Improve \\ the Generalization of Deep Neural Networks \\ on the Standard Supervised Learning? \thanks{The code will be released upon acceptance.}}

\author{\IEEEauthorblockN{Xiang Deng}
\IEEEauthorblockA{Computer Science Department\\
State University of New York at Binghamton \\
Email: xdeng7@binghamton.edu}
\and
\IEEEauthorblockN{Zhongfei Zhang}
\IEEEauthorblockA{Computer Science Department\\
State University of New York at Binghamton \\}
}


%


\maketitle

\begin{abstract}
Substantial efforts have been made on improving the generalization abilities of deep neural networks (DNNs) in order to obtain better performances without introducing more parameters. 
On the other hand, meta-learning approaches exhibit powerful generalization on new tasks in few-shot learning.
Intuitively, few-shot learning is more challenging than the standard supervised learning as each target class only has a very few or no training samples.
The natural question that arises is whether the meta-learning idea can be used for improving the generalization of DNNs on the standard supervised learning.
In this paper, we propose a novel meta-learning based training procedure (MLTP) for DNNs and demonstrate that the meta-learning idea can indeed improve the generalization abilities of DNNs.
MLTP simulates the meta-training process by considering a batch of training samples as a task.
The key idea is that the gradient descent step for improving the current task performance should also improve a new task performance, which is ignored by the current standard procedure for training neural networks. 
MLTP also benefits from all the existing training techniques such as dropout, weight decay, and batch normalization.
We evaluate MLTP by training a variety of small and large neural networks on three benchmark datasets, i.e., CIFAR-10, CIFAR-100, and Tiny ImageNet. 
The experimental results show a consistently improved generalization performance on all the DNNs with different sizes, which verifies the promise of MLTP and demonstrates that the meta-learning idea is indeed able to improve the generalization of DNNs on the standard supervised learning.
\end{abstract}


%
\IEEEpeerreviewmaketitle

\section{Introduction}
Deep neural networks (DNNs) have achieved state-of-the-art performances in varieties of applications in computer vision domain \cite{girshick2015fast} \cite{redmon2016you} \cite{liu2016ssd} .
It is often observed that larger neural networks exhibit a better generalization than smaller ones.
It is appealing to further tap the potential generalization abilities of DNNs, especially for small neural networks, expecting to secure better performances without introducing additional parameters.\par 


%

Many efforts have focused on improving the generalization abilities of DNNs. Model distillation based methods \cite{hinton2015distilling} \cite{ba2014deep} \cite{romero2014fitnets} \cite{parisotto2015actor} attempt to improve the generalization of student networks by transferring knowledge from teacher networks to them. Various regularizers, such as weight decay \cite{krogh1992simple} and the Jacobian matrix \cite{sokolic2017robust}, have been proposed to improve the generalization of DNNs by mitigating overfitting or enhancing stability. Other techniques like Dropout \cite{srivastava2014dropout} and Batch Normalization \cite{ioffe2015batch} which are directly implemented in the network architecture have also shown the ability to improve the performances of neural networks. 
\par

On the other hand, we notice that meta-learning approaches \cite{finn2017model} \cite{andrychowicz2016learning} have obtained a great success in few-shot learning with their excellent generalization abilities on new tasks.
Intuitively, few-shot learning is more challenging than the standard supervised learning as each target class only has a very few or no training samples.
In light of this, we ask whether the meta-learning idea can be used to improve the generalization of DNNs on the standard supervised learning.
To the best of our knowledge, this problem has not been explored in the existing literature, which is a novel and promising direction for improving the generalization abilities of DNNs.\par

In this paper, we develop a meta-learning based training procedure (MLTP) for training DNNs on the standard supervised learning and demonstrate that the meta-learning idea is able to improve the generalization of DNNs.
MLTP casts the standard training process to a meta-training process by considering a batch of samples as a task.
It explicitly requires that one gradient descent step on the current task also improves a new task performance, which is ignored in the current standard training process. 
Our work is most related to MAML \cite{finn2017model} which is a gradient based meta-learning approach for few-shot learning. 
MAML learns an appropriate initialization for a variety of tasks, and then fine-tunes a few gradient steps on a new task.
MLTP is different from MAML in many aspects due to their different problem settings.
First, the objective functions are different because of their different goals.
Second, MLTP does not require fine-tuning while MAML needs to fine-tune a few gradient steps on new tasks.
This is due to the fact that MLTP assumes training a model over a set of tasks in a domain and automatically generalizing on new similar tasks in the same domain.\par

The current standard procedure for training DNNs on supervised learning is to update parameters by iteratively doing gradient descent using the gradients on a batch of samples.
The major difference between MLTP and the standard training procedure is that MLTP trains neural networks through considering a batch of samples as a task and incorporating the meta-learning idea into the objective function. 
MLTP does not introduce any additional parameters to the neural network as it only changes the training process.
\par

In addition, we also provide more efficient variants of MLTP, i.e., $MLTP_{conv}$, $MLTP_{fc}$, and $MLTP_{FO}$.
We compare the performances of MLTP and its variants with those of the standard training procedure on different sizes of neural networks using three benchmark datasets, i.e., CIFAR-10 \cite{krizhevsky2009learning}, CIFAR-100 \cite{krizhevsky2009learning}, and Tiny ImageNet \footnote{http://tiny-imagenet.herokuapp.com/}.
Extensive experimental results show that MLTP and its alternative variants are able to improve the generalization abilities of neural networks with various sizes by only changing the training process without increasing the parameter numbers of the neural networks. This indicates that the meta-learning idea is able to improve the generalization of DNNs on the standard supervised learning.

The main contributions of our work can be summarized as follows:
\begin{itemize}
\item Considering the potential generalization capabilities of DNNs and the excellent generalization performances of meta-learning approaches on few-shot learning, we ask the question of whether the meta-learning idea is able to improve the generalization of DNNs on the standard supervised learning. We develop MLTP and its variants, and demonstrate that the meta-learning idea can indeed improve the generalization of DNNs.

\item  We evaluate MLTP and its variants using neural networks with different sizes on three benchmark datasets, i.e., CIFAR-10, CIFAR-100, and Tiny ImageNet. Experimental results have shown the superior performances of MLTP and its variants. We also provide the theoretical analysis of MLTP.

\item Different from the existing literature on improving the generalization of DNNs, to the best of our knowledge, we are the first to explore the meta-learning idea on this problem and obtain promising results, which gives a novel direction for the future work on this problem. 
We bridge the gap between meta-learning and the generalization of DNNs on the standard supervised learning through MLTP.

\end{itemize}

\section{Related Work}
Our work is related to the literature on explaining and improving the generalization of DNNs, as well as the gradient based meta-learning methods. Thus, we first review the existing methods about the generalization of DNNs, and then present the gradient based meta-learning approaches.

\subsection{Generalization of Deep Neural Networks}
Recently, substantial research has focused on explaining and improving the generalization abilities of DNNs.
On one hand, there are a number of efforts to explain the generalization of DNNs through the classification margin \cite{sokolic2017robust}, complexity \cite{neyshabur2017exploring}, sensitivity \cite{novak2018sensitivity}, Fourier analysis \cite{xu2018understanding} \cite{xu2018training} \cite{rahaman2018spectral}, information bottleneck \cite{shwartz2017opening} \cite{saxe2018information}, and empirical results \cite{zhang2016understanding}.
On the other hand, weight decay \cite{krogh1992simple} and dropout \cite{srivastava2014dropout} have been proposed to improve the generalization abilities of DNNs by mitigating overfitting.
Chaudhari et al. \cite{chaudhari2016entropy} develop a local-entropy-based objective function to avoid poorly-generalizable solutions, thus improving the generalization of DNNs.
Model distillation based methods \cite{hinton2015distilling} \cite{ba2014deep} \cite{romero2014fitnets} \cite{parisotto2015actor} attempt to improve the generalization performance of a neural network by transferring the knowledge from another network to it.
Deep mutual learning \cite{zhang2018deep} improves the generalization performance of DNNs by asking a set of networks to learn collaboratively and to teach each other.

\subsection{Gradient Based Meta-Learning}
Our work is most related to the gradient based meta-learning algorithms, especially MAML \cite{finn2017model} and MAML based meta-learning approaches. 
MAML \cite{finn2017model} learns an appropriate initialization on varieties of tasks, and then fine-tunes on new tasks by doing a few gradient descent steps with a very few samples.
Finn et al. \cite{finn2017meta} further explore the MAML idea and suggest that DNNs combined with the standard gradient descent be able to approximate any learning algorithm.
Yoon et al. \cite{kim2018bayesian} propose to learn the Bayesian posterior from a few-shot dataset through combining MAML with the nonparametric variational inference.
Auto-Meta \cite{kim2018auto} adopts the automated neural architecture search to MAML for finding the optimal architectures of meta-learners.
Reptile \cite{nichol2018first} which is closely related to first-order MAML (FOMAML) learns models for few-shot learning by a process like the joint training.
Grant et al. \cite{grant2018recasting} formulate MAML as a method for probabilistic inference in a hierarchical Bayesian model.
MLDG \cite{li2018learning} introduces MAML to domain generalization by modifying the objective function. 
Alpha MAML \cite{behl2019alpha} combines MAML with an online hyperparameter adaptation approach \cite{baydin2017online}, thus avoiding tuning the learning rates. 
MAML++ \cite{antoniou2018train} modifies MAML in several aspects and improves its stability and performances.
Jayathilaka et al. \cite{jayathilaka2019enhancing} develop a new technique called meta-step gradient pruning to improve the generalization of FOMAML and investigate the effects of increasing the network depth in FOMAML.

\section{Framework}
In this section, we first review the current standard procedure for training neural networks on supervised learning. Then we modify the standard training procedure through incorporating the meta-learning idea into the objective function, and present a novel meta-learning based training procedure (MLTP) which bridges the gap between meta-learning and neural network generalization on the standard supervised learning.

\subsection{Standard Procedure for Training DNNs on Supervised Learning}

Given training data $(X_{tra}, Y_{tra})$ where $X_{tra}$ are the inputs and $Y_{tra}$ are the targets, a neural network $f$ with parameters $w$ is trained on the data. In the standard training procedure, $w$ are updated iteratively based on the gradients calculated by using different batches of samples. Specifically, in each iteration, a batch of training samples $(x_{bat}, y_{bat})$ are utilized to compute the loss: 
\begin{equation}
\label{1}
C(w, x_{bat}, y_{bat}) = L(f(w, x_{bat}), y_{bat})
\end{equation}
where $L(.)$ is any loss function, such as mean square error and cross entropy, $w=[w_1, w_2, ..., w_n]$ with $w_i$ representing all the parameters in the $i$th layer of the neural network, and $n$ is the total number of layers in the neural network.
\par
With the loss from one batch of samples, the gradients with respect to $w$ can be calculated. Gradient descent is used to update $w$ iteratively based on the gradients on different batches of samples until the loss function converges. 

\subsection{Meta-learning Based Training Procedure}
Meta-learning approaches have shown excellent performances on few-shot learning with their powerful generalization abilities on new tasks. 
Inspired by this, we propose MLTP, a meta-learning based training approach for improving the generalization abilities of DNNs on the standard supervised learning. 
MLTP considers a batch of samples as a task and explicitly requires that the parameters $w$ after one gradient descent on the current task also work well on a new task. 
Thus, in every gradient descent iteration, MLTP randomly takes two different batches of training samples $(x_{bat}^i, y_{bat}^i)$ and $(x_{bat}^j, y_{bat}^j)$ as two tasks $task_i$ and $task_j$, respectively. 
The loss on the current task $task_i$ is written as:
\begin{equation}
\label{2}
C(w, x_{bat}^i, y_{bat}^i) = L(f(w, x_{bat}^i), y_{bat}^i)
\end{equation}

As MLTP requires the parameters $w$ after one gradient descent on the current task to also work well on a new task, the loss on the new task $task_j$ is written as:
\begin{equation}
\label{3}
\begin{split}
C(w - \alpha \frac{\partial L(f(w, x_{bat}^i),  y_{bat}^i)}{\partial w}, x_{bat}^j, y_{bat}^j)  = \\ L(f(w - \alpha \frac{\partial L(f(w, x_{bat}^i), y_{bat}^i)}{\partial w}, x_{bat}^j), y_{bat}^j)
\end{split}
\end{equation}
where $\alpha=[\alpha_1, \alpha_2, ... , \alpha_n]$ are online adapted variables with $\alpha_i$ denoting the inner step size shared by the $i$th layer's parameters $w_i$.

Therefore, the final objective function is the sum of the weighted losses from $task_i$ and $task_j$:

\begin{equation}
\begin{split}
\label{4}
J = C(w, x_{bat}^i, y_{bat}^i) + \\ \eta C(w- \alpha \frac{\partial L(f(w, x_{bat}^i), y_{bat}^i)}{\partial w}, x_{bat}^j, y_{bat}^j)
\end{split}
\end{equation}
where $\eta$ is a hyperparameter to balance the contributions between the two losses.\par

For the objective function (\ref{4}), it is worth noting that generally $\alpha$ are variables instead of hyperparameters. $\alpha$ are updated in the whole training process using gradient descent, which is different from the case on MAML \cite{finn2017model}. 
However, to reduce the computation for training large neural networks, one can elect to set $\alpha$ to a hyperparameter which can be tuned manually during the training process.
For both cases, MLTP trains DNNs by solving for the minimization problem with (\ref{4}) using a gradient descent based optimizer. 

\par

\begin{algorithm*} 
\caption{MLTP} 
 \begin{algorithmic} [1] 
 \renewcommand{\algorithmicrequire}{\textbf{Input:}} 
\renewcommand{\algorithmicensure}{\textbf{Output:}}
\REQUIRE Training data $(X_{tra}, Y_{tra})$, a neural network $f$ with parameters $w$
\ENSURE The optimal parameters $w$
\STATE \textbf{for} iterations = 1, 2, ..., n \textbf{do}
\STATE \qquad Randomly take two different batches of samples $(x_{bat}^i, y_{bat}^i)$ and $(x_{bat}^j, y_{bat}^j)$ as two tasks 
\STATE \qquad Compute the loss of the first task $(x_{bat}^i, y_{bat}^i)$: $ L(f(w, x_{bat}^i), y_{bat}^i)$
\STATE \qquad Do one gradient step to $w$: $w'= w - \alpha \frac{\partial L(f(w, x_{bat}^i), y_{bat}^i)}{\partial w}$ where $\alpha$ are the online adapted inner step sizes
\STATE \qquad Apply $w'$ to the second task $(x_{bat}^j, y_{bat}^j)$ to obtain the loss: $L(f(w-\alpha \frac{\partial L(f(w, x_{bat}^i), y_{bat}^i)}{\partial w}, x_{bat}^j), y_{bat}^j)$
\STATE \qquad  Obtain the final objective function: $J = L(f(w, x_{bat}^i), y_{bat}^i) + \eta L(f(w-\alpha \frac{\partial L(f(w, x_{bat}^i), y_{bat}^i)}{\partial w}, x_{bat}^j), y_{bat}^j)$
\STATE \qquad Update $w$ and $\alpha$:
$w = w -r \frac{\partial J}{\partial w }$; $\alpha = \alpha - r \frac{\partial J}{\partial \alpha }$ where $r$ is the learning rate
\STATE \textbf{end for}
\end{algorithmic} 
\end{algorithm*}

At the test time, MLTP does not need any additional computation compared with the standard training procedure because MLTP only changes the training process by incorporating the meta-learning objective. 
At the training time, minimizing (\ref{4}) needs the second derivatives with respect to $w$, which requires additional computation.
It is acceptable when the neural network is small or one's only concern is the test accuracy. 
For large neural networks, computing second derivatives increases more computation costs. 
To address this issue, we introduce a series of simplified alternative variants of MLTP at the end of this section. The experimental results show that the simplified variants of MLTP are able to work as well as MLTP, but with a much lower computational cost.\par

Our framework MLTP is summarized in Algorithms 1.

\subsection{Theoretical Analysis of MLTP}
In this section, we discuss why MLTP works and the differences between MLTP and the standard training procedure.\par

In MLTP, the parameters are updated by minimizing (\ref{4}). Intuitively, (\ref{4}) requires $w$ after one gradient step on the current task to also work well for a new task. To better understand the motivation of (\ref{4}), we provide the first-order Taylor expansion of (\ref{4}):

\begin{equation}
\begin{split}
\label{6}
J = C(w, x_{bat}^i, y_{bat}^i) + \eta C(w, x_{bat}^j, y_{bat}^j)-\\
\eta \alpha \frac{\partial C(w, x_{bat}^i, y_{bat}^i)}{\partial w} . \frac{\partial C(w, x_{bat}^j, y_{bat}^j)}{\partial w}
\end{split}
\end{equation}
where $.$ denotes the inner product operation. \par

The first two terms on the right hand side of (\ref{6}) mean minimizing the losses on both $task_i$ and $task_j$. The third term on the right hand side of (\ref{6}) is the negative inner product of the gradients with respect to $w$ on the two tasks. If the two gradients are in the same direction, the inner product is large and the negative inner product is small. Therefore, the third term tries to maximize the similarity between the gradients on the two tasks. This is also the main difference between MLTP and the standard training procedure, because if we take two batches of training samples in the standard training procedure to update $w$ in one step, the gradients are the average of the first two terms on the right hand side of (\ref{6}). But for MLTP, there is a third term to penalize the objective with the negative similarity between the two gradients.

\subsection{How to Implement the Widely Used Techniques in MLTP}
MLTP is a novel training procedure for the standard supervised learning. It benefits from all the existing training techniques, such as weight decay, batch normalization, and dropout. Specifically, for the regularization techniques like weight decay, we can implement them in MLTP by adding a regularization term in the final objective (\ref{4}):

\begin{equation}
\begin{split}
\label{5}
J_{Reg} = C(w, x_{bat}^i, y_{bat}^i) + \\ \eta C(w- \alpha \frac{\partial L(f(w, x_{bat}^i), y_{bat}^i)}{\partial w}, x_{bat}^j, y_{bat}^j) + \beta R(w)
\end{split}
\end{equation}
where $R(.)$ is the regularization on the parameters of DNNs and $\beta$ is the hyperparameter for this regularization. For example, $R(.)$ can be the L2 regularizer in the weight decay technique.\par

For the techniques like batch normalization and dropout, they are directly implemented in the network architecture $f$. Clearly MLTP can benefit from them by just directly minimizing (\ref{4}).

\subsection{Alternative MLTP Variants}
Minimizing (\ref{4}) requires the second derivatives with respect to $w$, which may be computationally expensive, especially for large neural networks, although it is acceptable in the cases where the neural network size is small or one's only concern is the test accuracy. 
To address this issue, we introduce the alternative MLTP variants. \par
It is well known that good features can improve model performances. Convolutional layers extract low-level and high-level features for the input data. Thus, we propose the first variant $MLTP_{conv}$ which only applies MLTP to the convolutional layers of a neural network. On the other hand, fully connected layers are used to combine the local features and generate the final predictions. Thus, the second variant of MLTP is $MLTP_{fc}$ which only applies MLTP to the fully connected layers. $MLTP_{conv}$ and $MLTP_{fc}$ are able to reduce the computational cost to different extents according to different neural network architectures. But for very large DNNs, the cost is still large. To address this issue, we introduce $MLTP_{FO}$ which only uses the first-order derivatives of (\ref{4}) to update $w$ by ignoring the second derivatives. $MLTP_{FO}$ has very little additional cost for calculating the derivatives with respect to $\alpha$ compared with that of the standard training procedure. Moreover, when $\alpha$ are set to a hyperparameter, there is almost no additional cost.
\par

In the above, we have proposed three alternative variants of MLTP, but it is worth noticing that MLTP is flexible. One can make his (her) own trade-off between the computational cost and the performance goal. For example, one may only apply MLTP to the first convolutional layer as the the first convolutional layer is closest to the input data while in another case one may elect to apply MLTP to the last fully connected layer considering that the last layer is closest to the output.



\begin{table*}[]
\centering
\caption{Test Accuracies (\%) of the Standard Training Procedure, MLTP, and Its Variants on CIFAR-10 }
\label{m1}
\begin{threeparttable}
\begin{tabular}{|c|c|cccc|}
\hline
\multirow{2}{*}{} & \multirow{2}{*}{Standard Training} & \multicolumn{4}{c|}{Ours}                    \\ \cline{3-6} 
                  &                                    & MLTP & $MLTP_{conv}$ & $MLTP_{fc}$ & $MLTP_{FO}$      \\ \hline
CNet1             & 81.9                               & 82.4 & 82.3       & 82.3     & \textbf{82.6} \\ \hline
CNet2             & 86.0                               & 86.4 & 86.3       & 86.4     & \textbf{86.7} \\ \hline
CNet3             & 85.9                               & 86.5 & 86.5       & 86.6     & \textbf{86.7} \\ \hline
CNet4             & 93.3                               & - \tnote{*}   & -\tnote{*}          & - \tnote{*}       & \textbf{93.6} \\ \hline
\end{tabular}
      \begin{tablenotes}
        \footnotesize
        \item[*] We only use $MLTP_{FO}$ to train CNet4 due to the high cost for computing the second derivative.
      \end{tablenotes}
  \end{threeparttable}
\end{table*}

\begin{figure*}[h!]
   \begin{minipage}{0.33\textwidth}
     \centering
     \includegraphics[width=.99\linewidth]{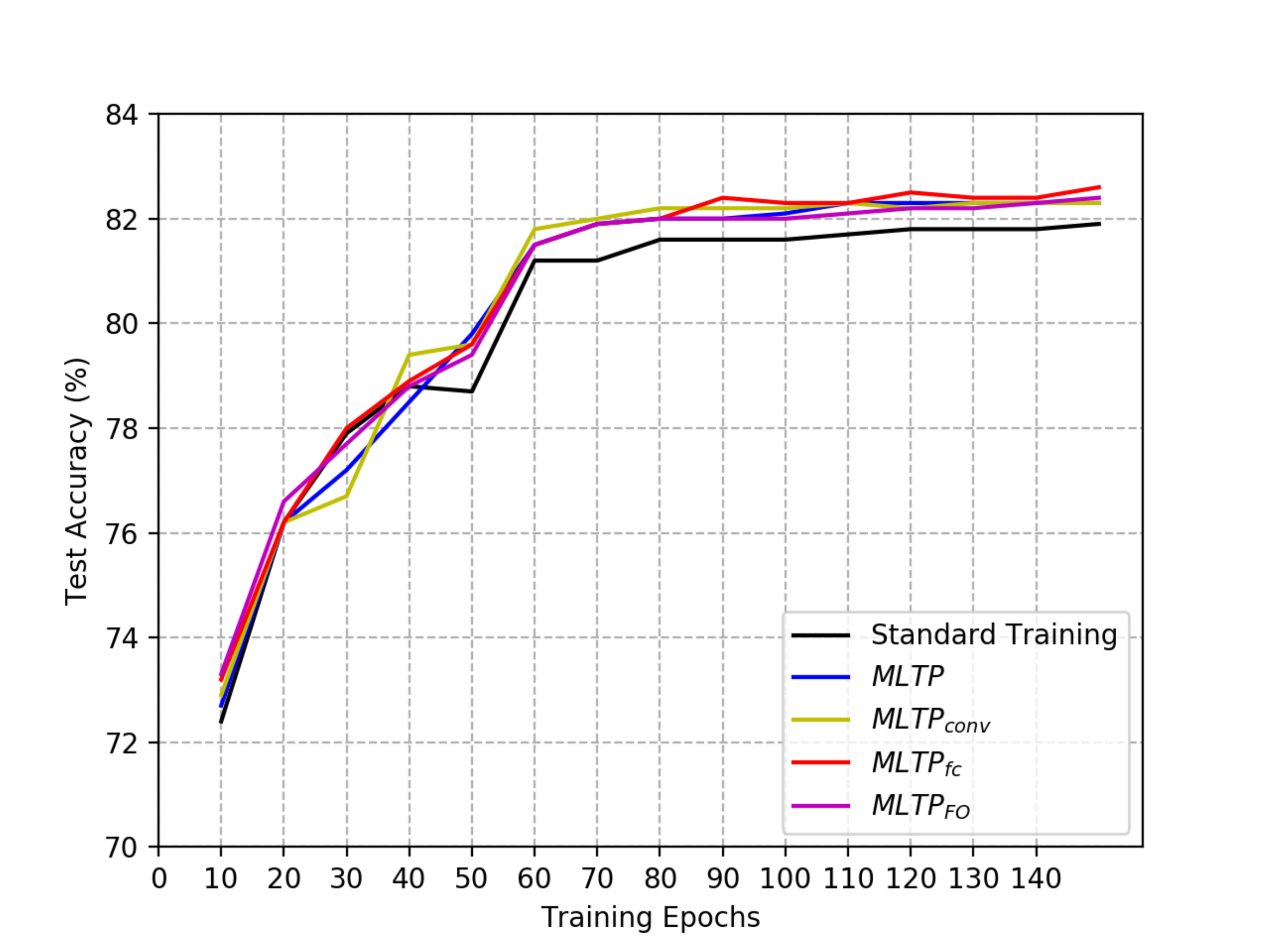}
     \caption{Training Curves of CNet1 on CIFAR-10}\label{f2}
   \end{minipage}\hfill
   \begin{minipage}{0.33\textwidth}
     \centering
     \includegraphics[width=.99\linewidth]{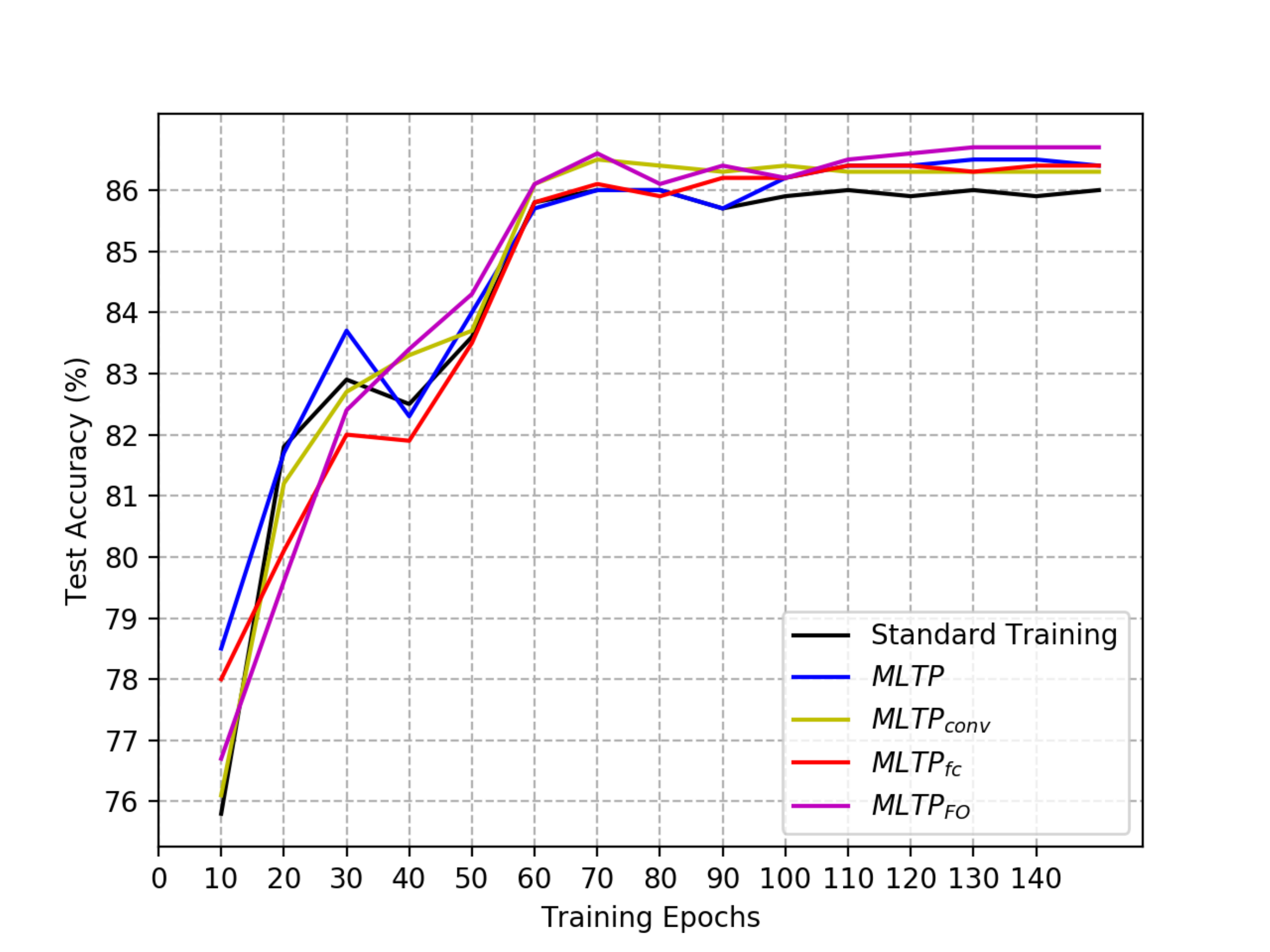}
     \caption{Training Curves of CNet2 on CIFAR-10}\label{f3}
   \end{minipage}\hfill
     \begin{minipage}{0.33\textwidth}
     \centering
     \includegraphics[width=0.99\linewidth]{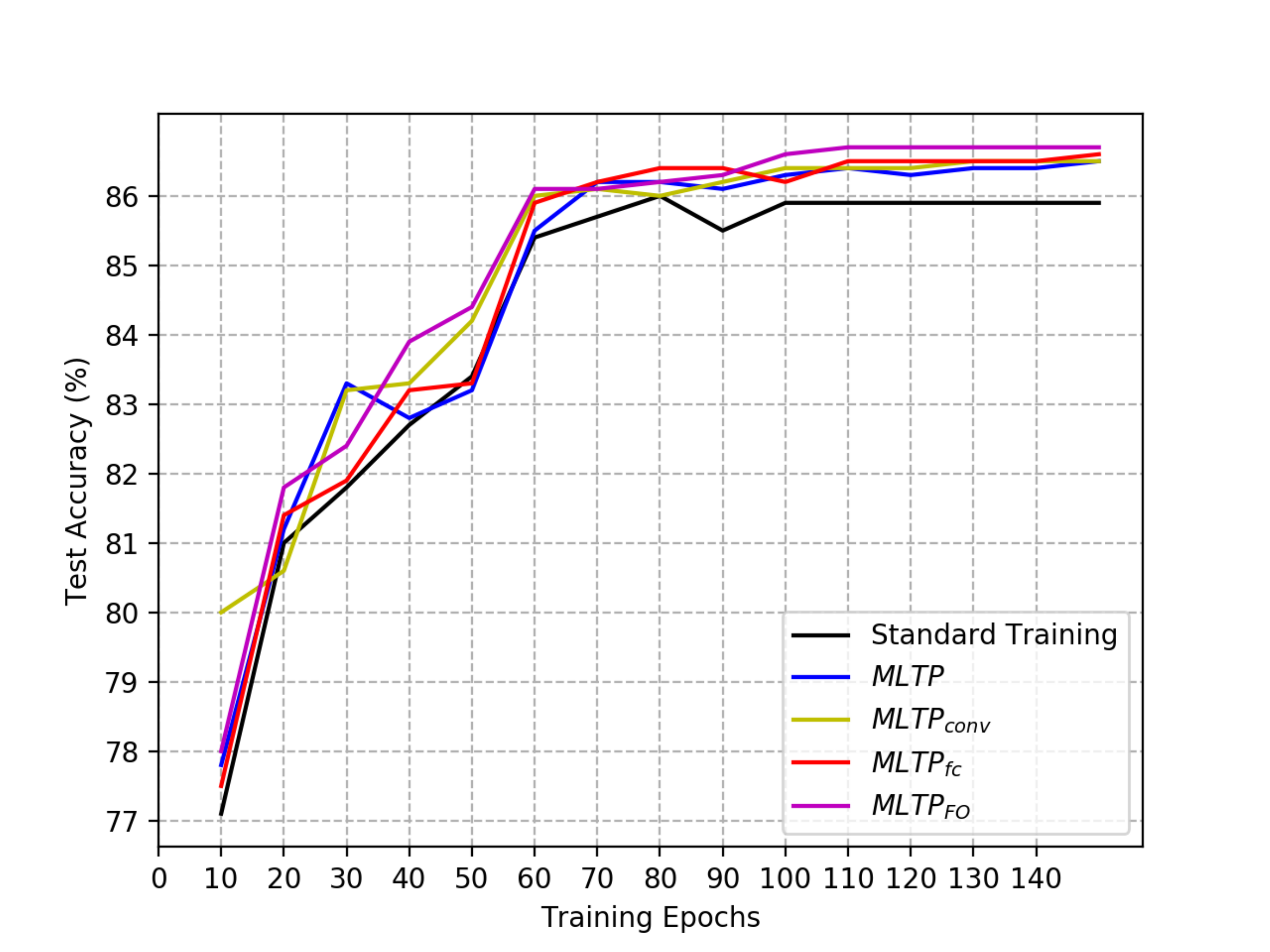}
     \caption{Training Curves of CNet3 on CIFAR-10}\label{f4}
   \end{minipage}
\end{figure*}

\section{Experiments}

We have introduced MLTP and its variants to connect between meta-learning and the standard supervised learning.
To verify that the meta-learning idea can improve the generalization of DNNs through MLTP and its variants, we compare the results of MLTP and its variants with those of the standard training procedure.
We report a series of experiments on neural networks with various sizes on three benchmark datasets, i.e., CIFAR-10 \cite{krizhevsky2009learning}, CIFAR-100 \cite{krizhevsky2009learning}, and Tiny ImageNet \footnote{http://tiny-imagenet.herokuapp.com/}. 
Through these experiments, we intend to address the following questions:
\begin{itemize}
\item Do \textbf{MLTP and its alternative variants all} yield better generalization performances for training neural networks than the standard training procedure?
\item Do MLTP and its alternative variants achieve consistently improved generalization performances on \textbf{neural networks with different sizes}?
\end{itemize}

It is also worth pointing out that the existing techniques, such as weight decay, dropout, and batch normalization, are not the competitors to MLTP, since these techniques can be implemented in MLTP and its variants easily.
To the contrary MLTP and its variants are able to benefit from them.
MLTP only changes the training process by introducing meta-training.
Thus, we only compare MLTP and its variants with the standard training procedure.
In every case below, the experiments are repeated three times and then we report the average test accuracy.

\par

To represent neural networks with different sizes used in the experiments, we define some notations in advance: \newline
(1) $Conv(c, f)$ represents the convolutional layer with $c$ kernels of size $f \times f$ \newline
(2) $Fc(m)$ represents the fully connected layer with output size $m$. \newline
(3) $MP$ is the max-pooling layer. \newline
(4) $Softmax(n)$ is the Softmax layer with output size n. \newline

\subsection{Results On CIFAR-10}
CIFAR-10 \cite{krizhevsky2009learning} is an image classification dataset with 10 classes, containing 50000 training images and 10000 test images with image size 32 $\times$ 32 in RGB space. We follow the standard preprocessing and data augmentation. Each image is preprocessed by subtracting its mean and dividing it by its standard deviation. During training time, we pad 4 pixels on each side of an image and randomly flip it horizontally. Then the image is randomly cropped to 32 $\times$ 32 size. 
During test time, we only evaluate the single view of an original 32 $\times$ 32 image without padding or cropping. \par

To demonstrate the effectiveness of MLTP and its variants, we use VGG \cite{simonyan2014very} inspired neural networks with different lengths and widths on CIFAR-10:\newline
(1) \textbf{CNet1}: $Conv(256, 3)$ + $MP$ + $Fc(512)$ \newline + $Softmax(10)$ \newline
(2) \textbf{CNet2}: $Conv(128, 3)$ + $Conv(128, 3)$ + $MP$ + \newline $Fc(256)$ + $Softmax(10)$ \newline
(3) \textbf{CNet3}:  $Conv(128, 3)$ + $Conv(128, 3)$ + $MP$ + \newline $Fc(256)$+ $Fc(256)$ + $Softmax(10)$ \newline
(4) \textbf{CNet4}:  $Conv(128, 3)$ + $Conv(128, 3)$ + $MP$ + \newline $Conv(256, 3)$ + $Conv(256, 3)$ + $MP$ + $Conv(512, 3)$ + \newline $Conv(512, 3)$ + $MP$ + $Fc(1024)$ + $Softmax(10)$; and weight decay with hyperparameter 1e-4 is used in CNet4. \par

We use the standard training procedure, MLTP, and its variants to train the four neural networks on CIFAR-10 and compare their performances. 
The batch size is set to 128. 
All weights are initialized with Xavier initializer \cite{glorot2010understanding}.
The initial learning rate is set to 0.001 and is divided by 10 after 50 and 100 epochs of training, and all networks have been trained for 150 epochs with the optimizer Adam \cite{kingma2014adam}.
$\alpha$ in MLTP and its variants are initialized with a normal distribution of mean 0.001 and standard deviation 0.001, and $\eta$ is simply set to 1.
Note that since MLTP needs two batches of training samples in each gradient descent iteration, to keep the batch size of MLTP the same as that of the standard training procedure, we split a batch of samples (128) into two smaller batches of samples (64) as two tasks in each iteration. The same strategy is adopted on CIFAR-100 and Tiny ImageNet. \par

\begin{figure*}[htb!]
   \begin{minipage}{0.33\textwidth}
     \centering
     \includegraphics[width=.99\linewidth]{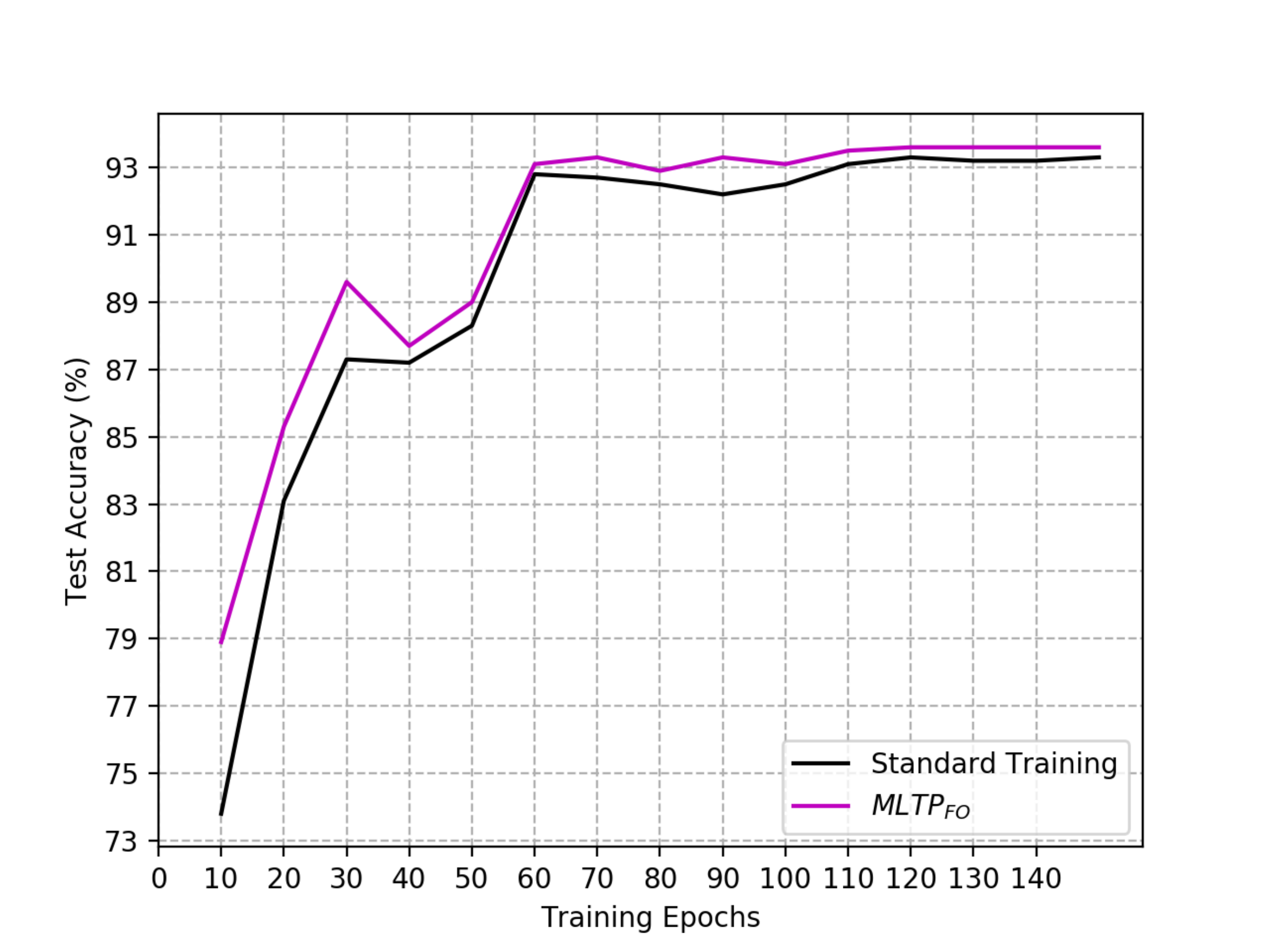}
     \caption{Training Curves of CNet4 on CIFAR-10}\label{f5}
   \end{minipage}\hfill
   \begin{minipage}{0.33\textwidth}
     \centering
     \includegraphics[width=.99\linewidth]{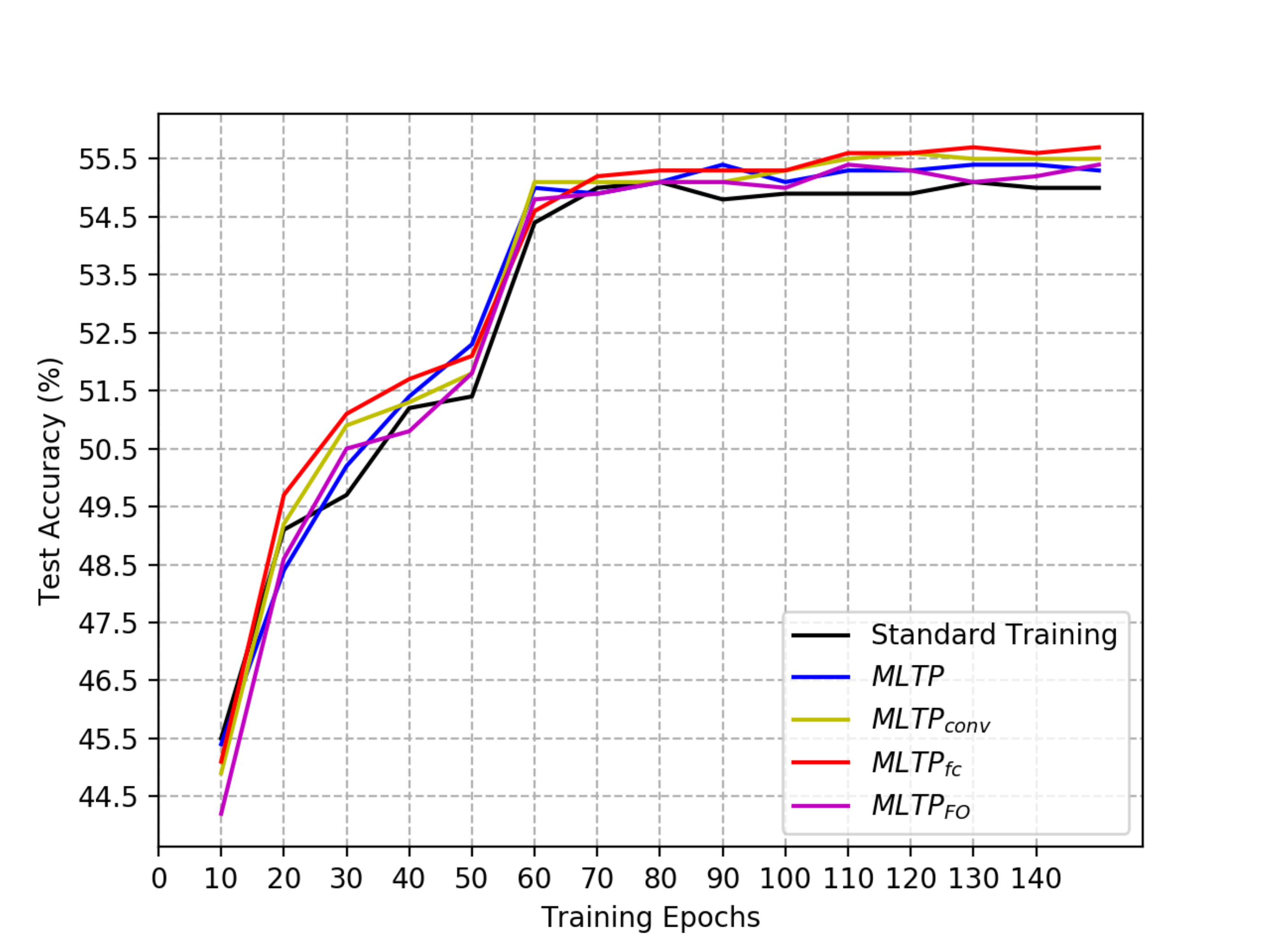}
     \caption{Training Curves of CCNet1 on CIFAR-100}\label{f6}
   \end{minipage}\hfill
   \begin{minipage}{0.33\textwidth}
     \centering
     \includegraphics[width=.99\linewidth]{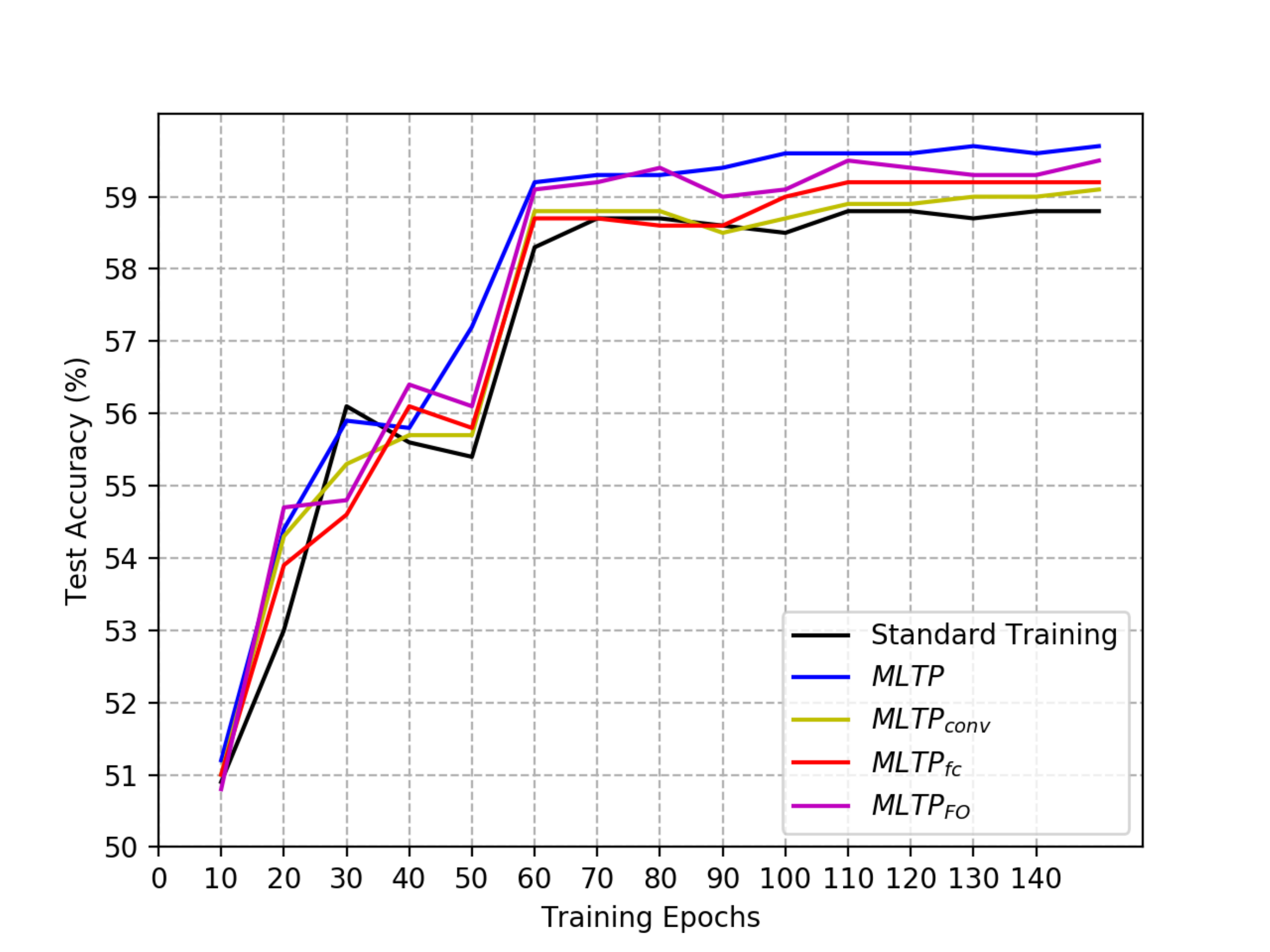}
     \caption{Training Curves of CCNet2 on CIFAR-100}\label{f7}
   \end{minipage}
\end{figure*}

Table \ref{m1} reports the comparison results on CIFAR-10. 
MLTP and its variants achieve similar results, and all outperform the standard training procedure on these four neural networks. 
The magnitudes of the improvements vary with the network architectures, but there is a consistent improvement on the generalization performance.
In addition, Figure \ref{f2}, Figure \ref{f3},  Figure \ref{f4}, and Figure \ref{f5} display the training curves of CNet1, CNet2, CNet3, and CNet4 with different training procedures, respectively. We observe that MLTP and its variants have the similar convergence speed to that of the standard procedure, but obtain better performances.

\par



\begin{table*}[]
\centering
\caption{Test Accuracies (\%) of the Standard Training Procedure, MLTP, and Its Variants on CIFAR-100 }
\label{m2}
\begin{threeparttable}
\begin{tabular}{|c|c|cccc|}
\hline
\multirow{2}{*}{} & \multirow{2}{*}{Standard Training} & \multicolumn{4}{c|}{Ours}                    \\ \cline{3-6} 
                  &                                    & MLTP & $MLTP_{conv}$ & $MLTP_{fc}$ & $MLTP_{FO}$      \\ \hline
CCNet1            &55.0            &55.3                & 55.5      &\textbf{55.7}    &55.4   \\ \hline
CCNet2            &58.8            & \textbf{59.7}      & 59.1      & 59.2            & 59.5\\ \hline
CCNet3           &58.4            & 58.5               & 59.0      & \textbf{59.5 }    & 59.0 \\ \hline
CCNet4            &71.9            & -\tnote{*}        & - \tnote{*}  &  -\tnote{*} &   \textbf{72.4} \\ \hline
\end{tabular}
      \begin{tablenotes}
        \footnotesize
        \item[*] We only use $MLTP_{FO}$ to train CNet4 due to the high cost for computing the second derivative.
      \end{tablenotes}
  \end{threeparttable}
\end{table*}

\begin{figure*}[htb!]
   \begin{minipage}{0.33\textwidth}
     \centering
     \includegraphics[width=.99\linewidth]{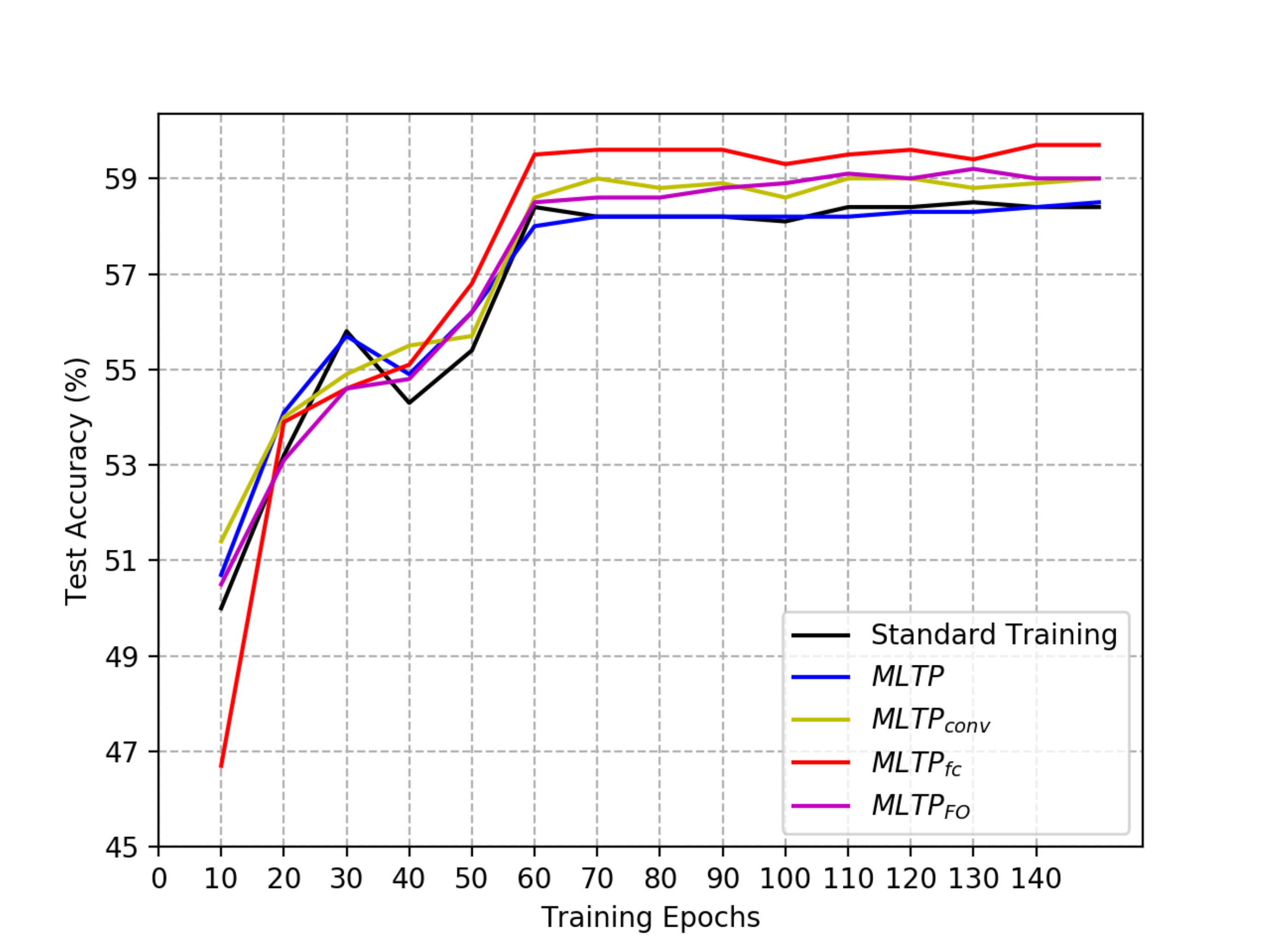}
     \caption{Training Curves of CCNet3 on CIFAR-100}
     \label{f8}
   \end{minipage}\hfill
   \begin{minipage}{0.33\textwidth}
     \centering
     \includegraphics[width=.99\linewidth]{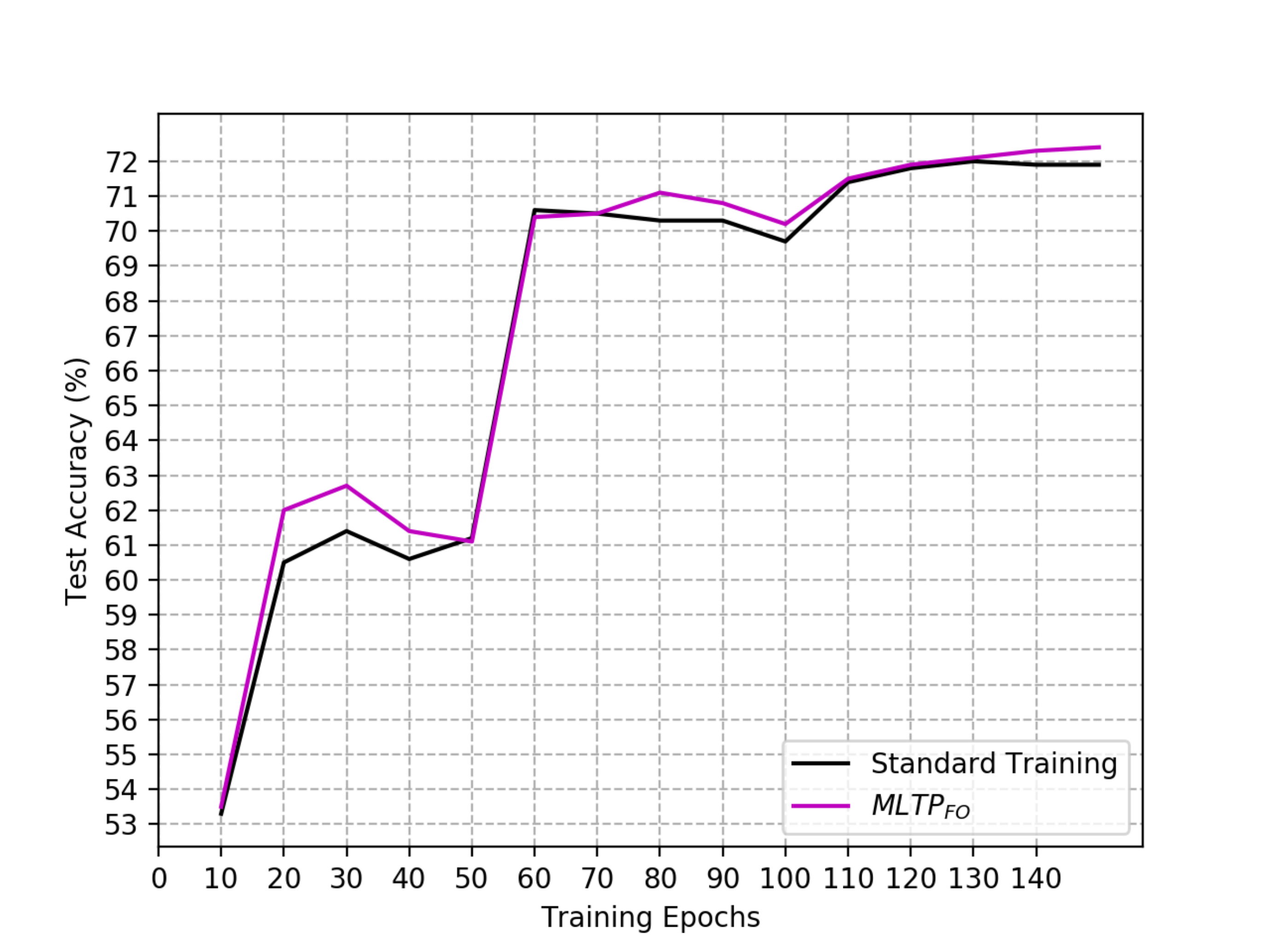}
     \caption{Training Curves of CCNet4 on CIFAR-100}
     \label{f9}
   \end{minipage}\hfill
   \begin{minipage}{0.33\textwidth}
    \centering
   \includegraphics[width=.99\linewidth]{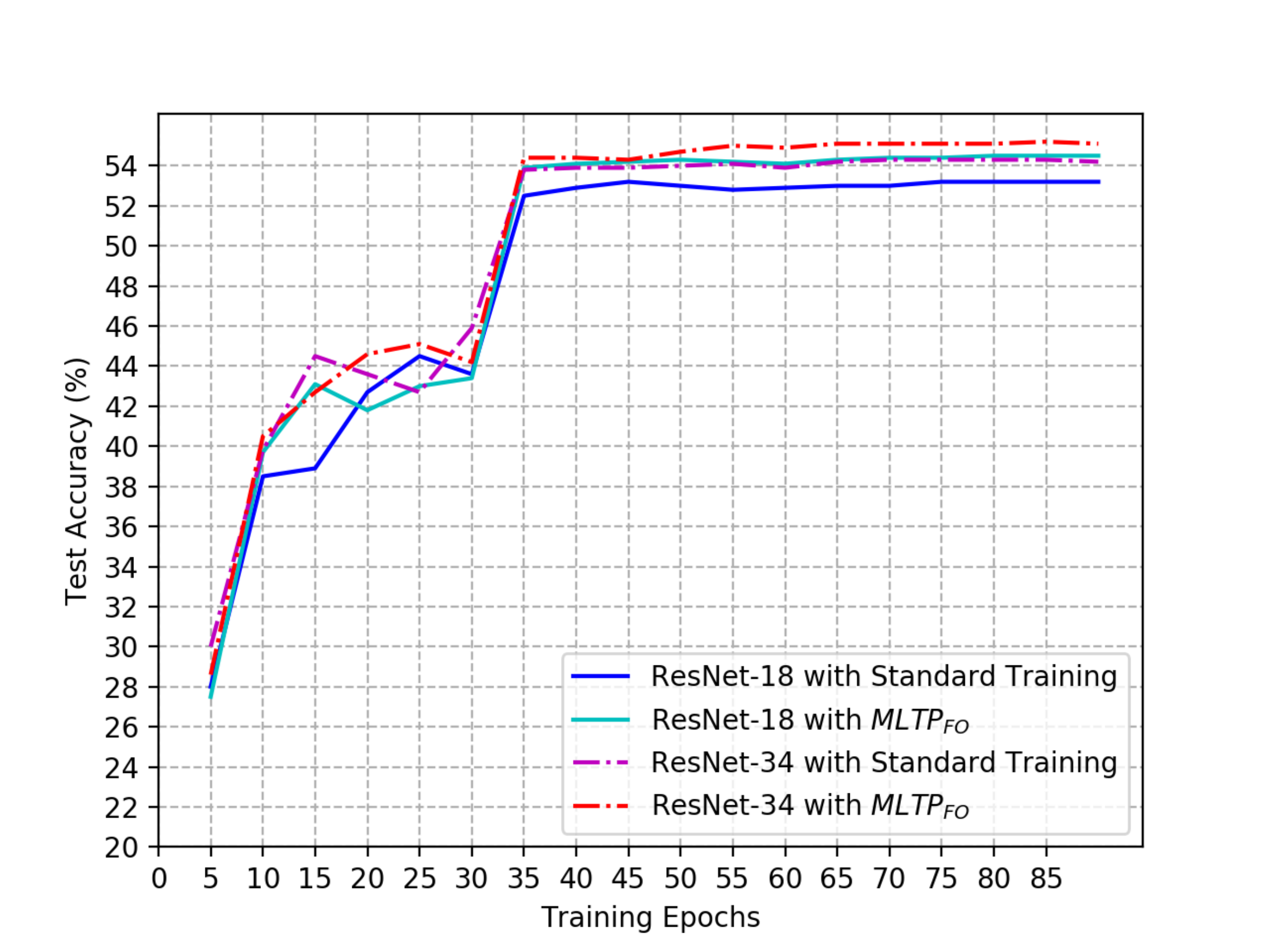}
   \caption[]{Training Curves of ResNet-18 and ResNet-34}
   \label{f10}
   \end{minipage}   
\end{figure*}

\subsection{Results On CIFAR-100}
CIFAR-100 \cite{lecun-mnisthandwrittendigit-2010} comprises similar images to those in CIFAR-10, but has 100 classes. We adopt the same preprocessing and data augmentation strategies as those in CIFAR-10. To demonstrate the promise of MLTP and its variants, we use different sizes of VGG inspired networks on CIFAR-100:\newline
(1) \textbf{CCNe1}: $Conv(256, 3)$ + $MP$ + $Fc(512)$ + \newline $Softmax(100)$\newline
(2) \textbf{CCNet2}: $Conv(128, 3)$ + $Conv(128, 3)$ + $MP$ + \newline $Fc(256)$ + $Softmax(100)$ \newline
(3) \textbf{CCNet3}: $Conv(128, 3)$ + $Conv(128, 3)$ + $MP$ + \newline $Fc(256)$ + $Fc(256)$ + $Softmax(100)$ \newline
(4) \textbf{CCNet4}:  $Conv(128, 3)$ + $Conv(128, 3)$ + $MP$ + \newline $Conv(256, 3)$ + $Conv(256, 3)$ + $MP$ + $Conv(512, 3)$ + \newline $Conv(512, 3)$ + $MP$ + $Fc(1024)$ + $Softmax(100)$; and weight decay with hyperparameter 1e-4 is used in CCNet4.\par

 
We use the different training procedures to train these networks on CIFAR-100 and compare their performances. 
The batch size is set to 128. 
All weights are initialized with Xavier initializer \cite{glorot2010understanding}.
The initial learning rate is 0.001 and is divided by 10 after 50 and 100 epochs of training.
The neural networks have been trained for 150 epochs with the optimizer Adam \cite{kingma2014adam}.
$\alpha$ in MLTP and its variants are initialized with a normal distribution of mean 0.001 and standard deviation 0.001 and $\eta$ is simply set to 1.\par

As seen from Table \ref{m2}, MLTP and its variants all obtain better performances than the standard training procedure on the four neural networks. 
This indicates that MLTP and its variants are able to improve the generalization performances of the neural networks with different sizes.
Moreover, as seen from Figure \ref{f6}, Figure \ref{f7}, Figure \ref{f8}, and Figure \ref{f9}, the convergence speeds of these training procedures are almost the same, but MLTP and its variants secure better generalization performances, which demonstrates the effectiveness of MLTP and its variants.
\par

\subsection{Results On Tiny ImageNet}
To further verify the promise of MLTP, we conduct several experiments with two large neural networks on Tiny ImageNet\footnote{http://tiny-imagenet.herokuapp.com/} i.e, a subset of ImageNet.
It is a challenging image classification dataset because each class only has a relatively small number of training samples.
Most prevalent networks fail to obtain promising results on it.
Specifically, it has 200 classes, each of which has 500 training and 50 validation images of size $64\times64$.
Each image is preprocessed by subtracting the mean of the whole training set and dividing it by the standard deviation.
At training time, we pad 8 pixels on each side of an image and randomly flip it horizontally. Then the image is randomly cropped to 64 $\times$ 64 size. At test time, we only evaluate the original image.

\par

The widely used ResNet-18 \cite{he2016deep} and ResNet-34 \cite{he2016deep} are adopted as the architectures.
We remove their first max-pooling layers as the input image size is already very small.
To save the cost, we only compare $MLTP_{FO}$ with the standard training procedure.
The batch size is set to 128 and the weights are initialized with Kaiming initialization \cite{he2015delving}.
Weight decay is adopted with hyperparameter 1e-5.
The initial learning rates are 0.05 and divided by 10 after 30 and 60 epochs of training.
All neural networks have been trained for 90 epochs using  SGD with momentum 0.9.
$\alpha$ in $MLTP_{FO}$ are initialized with a normal distribution of mean 1e-2 and standard deviation 1e-2, and $\eta$ is simply set to 0.5.


\par


Table \ref{m3} reports the results of different training procedures on Tiny ImageNet. $MLTP_{FO}$ performs better than the standard training procedure on both networks in terms of both TOP1 and TOP5 accuracies, which indicates that $MLTP_{FO}$ is applicable and useful to large neural networks.
Additionally, as shown in Figure \ref{f10}, $MLTP_{FO}$ has almost the same convergence speed as that of the standard procedure, but obtains better performances. \par


\begin{table}[htb]
\centering
\caption{TOP1 and TOP5 Test Accuracies (\%) of the Standard Training Procedure and $MLTP_{FO}$ with ResNet-18 and ResNet-34 on Tiny ImageNet }
\label{m3}
\begin{tabular}{c|cc|cc}
\hline
          & \multicolumn{2}{c|}{Standard Training} & \multicolumn{2}{c}{$MLTP_{FO}$} \\ \hline
          & TOP1               & TOP5              & TOP1          & TOP5          \\ \hline
ResNet-18 &53.2             & 76.5                  & \textbf{54.5}   &\textbf{77.2} \\ \hline
ResNet-34 &54.3                  &77.1              & \textbf{54.9}  &  \textbf{77.2}     \\ \hline
\end{tabular}
\end{table}

\section{Discussion}
MLTP and its variants bridge the gap between meta-learning and the standard supervised learning.
As seen from the experiments, they have shown a consistent generalization improvement on different sizes of neural networks, which demonstrates that the meta-learning idea is indeed able to improve the generalization of DNNs on the standard supervised learning.\par
Now we analyze their advantages and disadvantages. 
On one hand, they show a consistent generalization improvement on different sizes of neural networks, and do not bring any additional costs at test time compared with the standard training procedure as they only change the training process without changing the network architecture.
On the other hand, for MLTP, $MLTP_{conv}$, and $MLTP_{fc}$, they need additional costs during training although it is acceptable in the cases where the neural network is small or one's only concern is the test performance.
For $MLTP_{FO}$, it is able to improve the generalization performances of both small and large networks with very little even no additional costs during training.


\section{Conclusion}
Considering that meta-learning approaches have obtained excellent performances on few-shot learning with their powerful generalization abilities on new tasks, we ask the question of whether meta-learning idea can be used to further tap the potential generalization abilities of DNNs on the standard supervised learning. 
We have proposed a meta-learning based training procedure (MLTP) and have demonstrated that it is indeed true. 
MLTP considers a batch of samples as a task and introduces the meta-learning idea to the objective function by explicitly requiring the weights after one gradient step on the current task to generalize well on a new task.
We have also provided the theoretical analysis behind MLTP. 
Moreover, to reduce the computational costs, we have introduced several alternative variants of MLTP. 
Experimental results with neural networks of various sizes on three benchmark datasets, i.e., CIFAR-10, CIFAR-100, and Tiny ImageNet have demonstrated the effectiveness and promise of MLTP and its variants.
To the end, we bridge the gap between the meta-learning idea and the generalization of DNNs on the standard supervised learning by MLTP and demonstrate that the meta-learning idea is indeed able to improve the generalization of DNNs on the standard supervised learning.






\bibliographystyle{IEEEtran}
\bibliography{bare_conf.bib}
%



\end{document}